\documentclass[conference]{IEEEtran}
\IEEEoverridecommandlockouts

\usepackage{multirow}
\usepackage{algorithmic}
\usepackage{graphicx}
\usepackage[table]{xcolor}  
\usepackage{url}
\usepackage{booktabs}
\usepackage{subcaption}
\usepackage{hyperref}
\usepackage{amsmath}

\def\BibTeX{{\rm B\kern-.05em{\sc i\kern-.025em b}\kern-.08em
    T\kern-.1667em\lower.7ex\hbox{E}\kern-.125emX}}
\begin{document}

\title{HealthQ: Unveiling Questioning Capabilities of LLM Chains in Healthcare Conversations\\}


\author{
\IEEEauthorblockN{
Ziyu Wang\textsuperscript{*}\textsuperscript{1}, 
Hao Li\textsuperscript{*}\textsuperscript{2}, 
Di Huang\textsuperscript{3}, 
Hye-Sung Kim\textsuperscript{\textdagger}\textsuperscript{4}, 
Chae-Won Shin\textsuperscript{\textdagger}\textsuperscript{4}, 
Amir M. Rahmani\textsuperscript{1}
}
\IEEEauthorblockA{
\textsuperscript{1}University of California, Irvine, Irvine, CA, USA \\
\textsuperscript{2}Columbia University, New York, NY, USA \\
\textsuperscript{3}Washington University in St. Louis, St. Louis, MO, USA \\
\textsuperscript{4}Kookmin University, Seoul, South Korea \\
\textsuperscript{*}Equal contribution. Co-first authors. \quad \textsuperscript{\textdagger}Equal contribution.\\
\{ziyuw31, amirr1\}@uci.edu, hl3776@columbia.edu, di.huang@wustl.edu, \{khs0020, mirageciel\}@kookmin.ac.kr
}
}

\maketitle

\begin{abstract}
Effective patient care in digital healthcare requires large language models (LLMs) that not only answer questions but also actively gather critical information through well-crafted inquiries. This paper introduces HealthQ, a novel framework for evaluating the questioning capabilities of LLM healthcare chains. By implementing advanced LLM chains, including Retrieval-Augmented Generation (RAG), Chain of Thought (CoT), and reflective chains, HealthQ assesses how effectively these chains elicit comprehensive and relevant patient information. To achieve this, we integrate an LLM judge to evaluate generated questions across metrics such as specificity, relevance, and usefulness, while aligning these evaluations with traditional Natural Language Processing (NLP) metrics like ROUGE and Named Entity Recognition (NER)-based set comparisons. We validate HealthQ using two custom datasets constructed from public medical datasets, ChatDoctor and MTS-Dialog, and demonstrate its robustness across multiple LLM judge models, including GPT-3.5, GPT-4, and Claude. Our contributions are threefold: we present the first systematic framework for assessing questioning capabilities in healthcare conversations, establish a model-agnostic evaluation methodology, and provide empirical evidence linking high-quality questions to improved patient information elicitation.
\end{abstract}

\maketitle

\section{Introduction}
\label{sec:introduction}

The integration of Artificial Intelligence (AI) in healthcare has transformed personal healthcare, enabling advancements in diagnostic accuracy~\cite{behari2024decision}, treatment personalization~\cite{yang2024chatdiet, yangtransforming}, and patient management~\cite{Wang_EMBC2024, wang2020guardhealth, cheng2024efflex, zhang2018medical}. At the core of this transformation are Large Language Models (LLMs), which exhibit exceptional abilities in understanding and generating human-like text. These capabilities have been harnessed in healthcare applications such as virtual assistants, automated diagnostic tools, and clinical reasoning systems~\cite{abbasian2023conversational}. Much of the existing research has focused on enhancing LLM-based question-answering (QA) capabilities, which are pivotal for delivering timely and accurate responses to patient inquiries. Datasets and benchmarks such as MedQA~\cite{zhang2018medical}, emrQA~\cite{pampari2018emrqa}, and MedMCQA~\cite{pal2022medmcqa} have been developed to train and evaluate the QA abilities of LLMs in healthcare.

Recent advancements have demonstrated that LLMs can be organized into structured workflows, referred to as \textbf{\textit{LLM healthcare chains}}, to address complex interactions~\cite{abbasian2023conversational, limediq, liu2024medchain, qiu2024llm}, These chains represent a structured sequence of operations designed to manage multi-step tasks in medical contexts, integrating LLMs with additional tools, external knowledge bases, and reasoning mechanisms. Unlike standalone models, LLM healthcare chains leverage components such as Retrieval-Augmented Generation (RAG)\cite{lewis2020retrieval} for retrieving domain-specific information, Chain of Thought (CoT) reasoning\cite{wei2022chain} for step-by-step logical processing, and reflective chains for iterative refinement of outputs. In the context of healthcare, LLM healthcare chains encompass workflows tailored to retrieve relevant medical knowledge, interpret patient inputs, and generate diagnostic or follow-up questions. For example, when a patient presents symptoms, the chain might retrieve pertinent medical data, analyze patient responses to identify gaps, and generate contextually appropriate questions to gather additional information. These chains facilitate dynamic, multi-turn interactions, ensuring that conversations are not only relevant but also contextually rich, making them particularly well-suited for applications in diagnostic reasoning, patient education, and treatment planning. This structured and adaptive approach distinguishes LLM healthcare chains from static single-turn question-answering systems, providing a robust framework for managing complex healthcare conversations.

Despite the progress in QA systems, effective patient care demands more than the ability to answer questions. Healthcare interactions often require LLM healthcare chains to proactively ask pertinent and context-aware questions, eliciting critical information such as symptoms, medical history, and lifestyle factors. This questioning capability is crucial for accurate diagnosis and personalized treatment, yet it remains significantly under-explored~\cite{wu2023autogen, yang2023harnessing}. Existing benchmarks, which primarily evaluate static, single-turn QA performance, are insufficient for assessing the interactive and information-seeking abilities of LLMs. The MediQ~\cite{limediq} benchmark highlights the importance of transitioning from static to interactive paradigms, where LLMs must actively gather missing details via follow-up questions to achieve reliable clinical reasoning. These insights motivate the need for a dedicated framework to evaluate the question-asking capabilities of LLM healthcare chains in realistic settings.

To address this gap, we propose \textbf{HealthQ}, a novel framework for evaluating the questioning capabilities of LLM healthcare chains. HealthQ assesses both the quality of questions generated during interactions and their effectiveness in eliciting informative responses. Our framework employs a multi-dimensional evaluation methodology, incorporating traditional Natural Language Processing (NLP) metrics (e.g., ROUGE and Named Entity Recognition), as well as LLM-based interrogation metrics (e.g., relevance, specificity, and specificity). We validate the framework using custom datasets derived from public medical note datasets such as ChatDoctor and MTS-Dialog, ensuring realistic and comprehensive evaluation scenarios.

Our contributions are threefold:
\begin{enumerate}
    \item We conduct the first comprehensive study on the questioning capabilities of LLM healthcare chains, validating performance across multiple LLMs (GPT-3.5, GPT-4, Claude) and establishing metrics for evaluating question quality in healthcare contexts.
    \item We develop a dataset generation pipeline with validated parsing processes, enabling systematic evaluation of question-asking capabilities in interactive patient interactions.
    \item We propose an evaluation framework that integrates multiple LLM chain architectures and NLP metrics, demonstrating consistent performance across different LLM judges and providing reproducible benchmarks for medical question generation.
\end{enumerate}

\textit{The code, dataset generation pipeline, and evaluation framework are publicly available at 
\textcolor{blue}{\href{https://github.com/wangziyu99/HealthQ-LLM-Healthcare-Benchmark}{https://github.com/wangziyu99/HealthQ-LLM-Healthcare-Benchmark}}.}

\section{Related Work}
\label{sec:related_work}

\subsection{Digital Healthcare Interventions}
The advent of digital healthcare interventions has revolutionized patient care by leveraging technologies such as wearable devices~\cite{alikhani2024seal, alikhani2024ea, yao2020privacy, yang2024integrating, yang2023loneliness}, and AI-driven analytics~\cite{abbasian2023conversational, aqajari2024enhancing}. These interventions enable remote monitoring, personalized treatment plans, and predictive analytics, improving patient outcomes and reducing healthcare costs \cite{rahmani2018exploiting, wang2024ecg}. Novel AI integration, including machine learning and natural language processing, enhances diagnostic accuracy and patient management \cite{khatibi2024alcm, Wang_EMBC2024}. Machine learning algorithms predict disease progression and personalize treatments, while natural language processing interprets unstructured clinical data, enabling efficient information retrieval and decision support~\cite{sehanobishscalable}. These advancements highlight the potential of digital healthcare to improve the efficiency and effectiveness of care delivery.

\subsection{LLMs in Healthcare}
LLMs such as GPT-4 have shown remarkable proficiency in generating human-like text and understanding complex language constructs~\cite{achiam2023gpt, zhao2024bandit, li2025skewed}. These capabilities have been harnessed in various healthcare applications, including automated medical record summarization, virtual health assistants, and clinical decision support systems \cite{nori2023capabilities, yang2022large, zhao2024towards}. Automated medical record summarization condenses patient records into concise summaries, saving time for healthcare providers. Virtual health assistants interact with patients using natural language, providing answers to medical queries and reminders for medication, thereby enhancing patient engagement. Clinical decision support systems leverage the extensive knowledge embedded in LLMs to assist in diagnosing diseases and suggesting treatment options. LLMs streamline clinical workflows and improve patient engagement by providing accurate and timely responses to medical queries, ultimately enhancing the overall quality of care.

\subsection{Interactive Questioning with LLM Healthcare Chains}

Interactive healthcare conversational systems are crucial in providing patients with accurate and relevant information. Datasets and benchmarks like MedQA~\cite{jin2021disease} and MedMCQA~\cite{pal2022medmcqa} have played a significant role in advancing QA systems within the medical domain, focusing on optimizing the accuracy and relevance of answers generated by LLMs. These systems use large datasets to fine-tune model responses, making them contextually relevant and patient-centered. However, while answer generation has been prioritized, strategies for optimizing question formulation remain largely unexplored.

Several approaches have been introduced to refine the questioning capabilities of LLMs. Techniques such as Chain of Thought (CoT)~\cite{wei2022chain} prompting, reflective chains~\cite{asai2023self}, and RAG~\cite{lewis2020retrieval} integrate reasoning processes that help generate more contextually relevant questions. CoT prompting allows LLMs to break down complex queries into intermediate steps, leading to more precise and targeted questioning. Reflective chains enable the model to iteratively improve the questions by reflecting on previous intermediate answers, thus enhancing the relevance of subsequent queries. RAG, on the other hand, combines retrieval and generation processes to ensure that questions are informed by the most relevant information retrieved from a knowledge base. Despite these advancements, most existing work has primarily focused on optimizing task completion and response accuracy, with less attention given to the development of sophisticated questioning strategies that can drive more informative patient interactions. Our work addresses this gap by specifically evaluating the questioning capabilities of LLM healthcare chains, contributing a novel framework that benchmarks their ability to engage in meaningful and effective patient dialogues.

\section{Method}
\subsection{Overview of HealthQ}

For benchmarking, we customized several state-of-the-art LLM healthcare chains to simulate a doctor's role, using training data for search and retrieval without prior knowledge of test patients. Virtual patients were simulated using test data, and the LLM chains' generated questions were assessed. The evaluation framework includes generating patient statements, scoring questions, and comparing results with traditional NLP metrics. HealthQ uses two primary evaluation metrics: LLM judge interrogation and summarization-based evaluation. LLM judge interrogation involves an additional LLM evaluating question quality based on relevance, specificity, and informativeness. Summarization-based evaluation assesses how well-generated answers capture the complete patient case by comparing them to detailed patient information from the test data.

Figure~\ref{fig:system_framework} illustrates our system framework. Original medical notes are split into train and test sets. The train set builds a vectorized knowledge base. The test set, parsed by an LLM parser, produces ground truth pairs. The LLM healthcare chain generates questions using additional tools. A virtual patient simulator provides partial patient statements and answers. The generated questions are evaluated by an LLM judge based on various metrics. Finally, answers are assessed using summary-based metrics like NER2NER and ROUGE.

We also analyze the relationship between question quality and answer comprehensiveness through correlation and mutual information studies, demonstrating how well-formulated questions enhance patient information gathering and diagnostic accuracy.

\subsection{Data}

HealthQ necessitates a general data processing mechanism due to the unstructured nature of medical notes and the often ambiguous doctor-patient conversations. Our data reformatting serves two critical purposes: (1) establishing ground truth for evaluating generated questions by identifying what information doctors sought in successful consultations, and (2) creating separate knowledge bases for virtual patient simulation and question evaluation. When doctors ask effective questions in real consultations, they uncover specific patient information - this relationship between questions and discovered information forms our evaluation basis.

To accurately evaluate the questioning capabilities of LLM healthcare chains, we extract two key components from each consultation: the patient's known information (symptoms, feelings, and history) and the doctor's information-seeking behavior (what information the doctor aimed to uncover through their questions). This paired format allows us to assess whether generated questions would effectively uncover similar critical information. For example, if the original consultation shows a doctor successfully uncovering fever details through specific questions, we can evaluate whether our LLM chains generate questions that would similarly reveal this important information. To illustrate this relationship between original consultations and our evaluation framework, Table~\ref{tab:data_sample} shows an example of the original and processed data formats.

\subsubsection{Dataset}
To evaluate the questioning capabilities of LLM healthcare chains, we utilized two public datasets: ChatDoctor~\cite{li2023chatdoctor} and MTS-Dialog~\cite{benabacha2023mts}. The ChatDoctor dataset comprises 110,000 anonymized and grammatically corrected medical conversations, covering a wide range of symptoms, diagnoses, and treatment plans. This dataset, originally used to fine-tune a medical chat model based on LLaMA~\cite{touvron2023llama}, provides a robust foundation for assessing the LLM chains' ability to generate relevant and context-aware questions. The MTS-Dialog dataset includes 1,700 doctor-patient dialogues and corresponding clinical notes, capturing detailed exchanges across various medical specialties. These clinical notes summarize patient encounters, highlighting key medical facts, diagnoses, and treatment plans. The comprehensive nature of this dataset allows for a thorough assessment of the LLM chains' questioning capabilities in diverse medical scenarios. By utilizing these two datasets, we ensure a diverse and representative evaluation, covering a broad spectrum of medical conditions and conversational contexts.

\begin{table}[ht]
    \centering
    \caption{\textbf{Example of original and processed data.} The original format includes detailed patient background, symptoms, diagnosis, history, plan of action, and dialogue. The processed format extracts essential patient known knowledge and reformulates doctor question statements for clarity and consistency.}
    \label{tab:data_sample}
    \scriptsize 
    \begin{tabular}{|p{8cm}|}
    \hline
    \rowcolor[HTML]{EFEFEF} 
    \textbf{Original Data Format} \\
    \hline
    \textbf{Background:} \\
    \textbf{Symptoms:} Congestion, increased coughing, fever (temperature of 101°F). \\
    \textbf{Diagnosis:} Upper respiratory infection, possible pertussis. \\
    \textbf{History of Patient:} Patient experiencing congestion, increased coughing, and fever; possibility of pertussis indicated by doctor, inquiring about presence of apnea. \\
    \textbf{Plan of Action:} Doctor to \colorbox{green}{evaluate and determine} appropriate course of action. \\
    \textbf{Dialogue:} \\
    Doctor: How's he feeling today? \\
    Guest\_family: I think this is the worst he's been feeling all week. \\
    Doctor: I'm sorry to hear that he hasn't been feeling well. What symptoms has he been having? \\
    Guest\_family: He's been very \colorbox{red}{congested} as of late and seems to be \colorbox{red}{coughing} a lot more than usual. He was also running a \colorbox{red}{fever} yesterday. \\
    Doctor: I see. What was his temperature? \\
    Guest\_family: The thermometer was reading one hundred and one degrees Fahrenheit. Does he need to go to the hospital? \\
    Doctor: Let me evaluate and see what we can do for him today. \\
    Guest\_family: Thank you, doctor. \\
    Doctor: I suspect that he has an \colorbox{pink}{upper respiratory infection}, possible \colorbox{pink}{pertussis}. Is he still experiencing \colorbox{orange}{apnea}? \\
    \hline
    \rowcolor[HTML]{EFEFEF} 
    \textbf{Processed Data Format} \\
    \hline
    \textbf{Patient Known Knowledge:} \\
    The patient has been very \colorbox{red}{congested} lately and \colorbox{red}{coughing} more than usual. He had a \colorbox{red}{fever} of 101°F yesterday. \\
    \textbf{Doctor Question Statements:} \\
    The doctor asked about the patient's symptoms and temperature. The doctor suspects the patient has an \colorbox{pink}{upper respiratory infection}, possibly \colorbox{pink}{pertussis}, and asked if the patient is still experiencing \colorbox{orange}{apnea}. The doctor said he will \colorbox{green}{evaluate and see} what they can do for the patient today. \\
    \hline
    \end{tabular}
\end{table}

\begin{figure*}[ht]
    \centering
    \includegraphics[width=\linewidth]{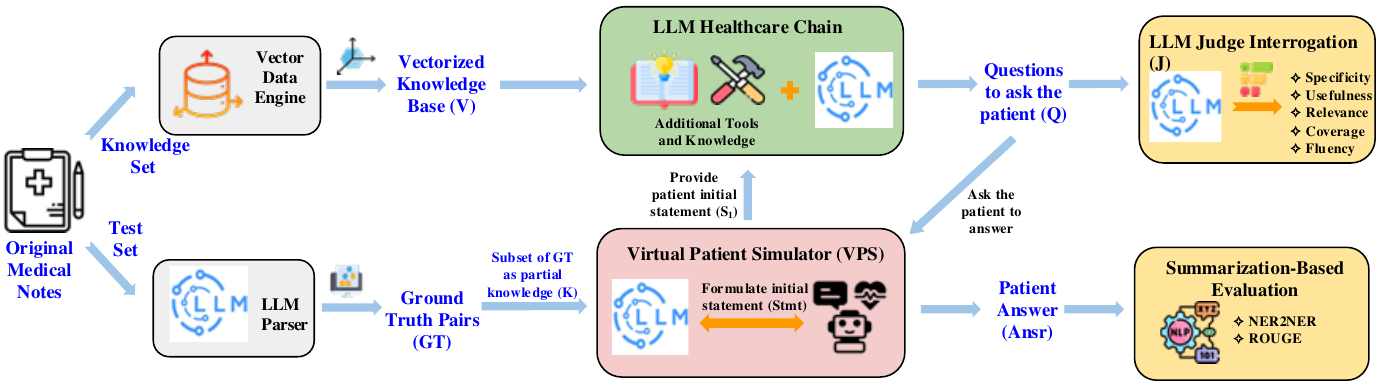}
    \caption{System framework for evaluating the questioning capabilities of LLM healthcare chains.}
    \label{fig:system_framework}
    \vspace{-0em} 
\end{figure*}

\subsubsection{Data Processing}
As depicted in Figure~\ref{fig:system_framework}, we partition the medical notes into two sets: one for constructing the knowledge base and another for evaluation. The first set is indexed in a vector database (VDB), serving as a reference knowledge base that LLM chains can query during operation - similar to how RAG systems typically operate. This is not traditional model training, but rather creation of a searchable medical knowledge repository. The evaluation set remains separate to assess the chains' questioning capabilities on entirely new patient scenarios.

Specifically, we construct the knowledge base using FAISS\footnote{\url{https://github.com/facebookresearch/faiss}} for efficient similarity search. The customized collate functions specify comparison fields and utilize metadata for enhanced information retrieval. This flexibility accommodates diverse datasets with varying forms and topics. During operation, multiple VDB searching functions can be provided as tools without merging, allowing the LLM to select the appropriate knowledge source. VDBs work by finding the nearest vectors, which are represented by embedding models. Each vector database entry is defined as: 

\[
V_i = [\text{Record}_i, \mathbf{v}_i], 
\]

where \( \mathbf{v}_i \) is the vector representation used for comparison to find the nearest vectors in the VDB for efficient medical knowledge retrieval.

Each entry in the vector database, referred to as a record, is structured to include both the content and the associated metadata. The collate function processes the input data to create these records:

\[
\text{Record}_i = (\text{Content}_i, \text{Metadata}_i) = \text{Collate}(\text{Data}_{\text{kb}}(i))
\]

This knowledge base facilitates efficient retrieval and processing of medical information during question generation. For the evaluation set, we employed Claude\footnote{\url{https://claude.ai/}} to parse patient notes into structured formats, producing pairs of patient known knowledge and doctor question statements. The final evaluation ground truth, \(G\), is thus composed of:

\[
GT = \{\text{LLM\_Parser}(
\text{Data}_{\text{eval}}(i))\}
\]

\subsection{Baseline Healthcare LLM Chains}

Healthcare LLM chains are structured workflows that utilize LLMs to simulate patient-doctor interactions by generating diagnostic questions and interpreting responses. These chains serve as frameworks for improving the relevance, specificity, and informativeness of questions in healthcare conversations, enabling precise information gathering. In this work, we implement several state-of-the-art LLM chains, each designed with unique mechanisms for question generation.

\textbf{Hardcoded Workflow:} This is the lightest chain that does not require LLM inference but only uses LLM embedding and NER in a pre-defined workflow. Initially, it takes in a patient's first statement and retrieves relevant cases from the vector database using a search function. NER is applied to extract symptoms and medications from the retrieved cases. It then confirms each extracted symptom and medication with the patient using predefined templates and updates the state accordingly.

\textbf{RAG ~\cite{lewis2020retrieval}:} This chain integrates retrieval and generation processes. It uses a retrieval mechanism to gather relevant cases from the vector database and then employs a language model to generate questions. The questions are designed to be concise and specific, leveraging the retrieved information to enhance their relevance and utility. We compare two approaches: one using an existing framework, LangChain RetrievalQA\footnote{\url{https://docs.smith.langchain.com/old/cookbook/hub-examples/retrieval-qa chain}}, and another customized version tailored specifically for generating questions rather than answers, aligning with the goals of our evaluation framework.

\textbf{RAG with Reflection~\cite{asai2023self}:} Extending the basic RAG framework, this chain includes a reflection mechanism. After generating an initial question, the chain reflects on the quality and relevance of the question by re-evaluating the context of the retrieved cases. This reflective process allows the chain to refine and improve the question, making it more targeted and contextually appropriate.

\textbf{RAG with CoT~\cite{wei2022chain}:} This chain incorporates a CoT approach, involving iterative reasoning steps. It generates an initial question(or intermediate thinkings) and then performs several iterations of reasoning to refine the question. Each iteration considers additional contextual information from the retrieved cases, progressively enhancing the question's specificity and relevance.

\textbf{RAG with Reflection and CoT-Self-Consistency (CoT-SC)~\cite{wang2022self}:} Building on the CoT approach, this chain introduces self-consistency checks. It branches several CoT from the same query, and gather the results of which to see of the different branches reaches similar results. This ensures that the final question is not only refined but also consistent and robust, reducing the likelihood of contradictions. Different from the original CoT-SC which uses a voting approach for the final result, our version uses LLM to compare the several branches of the final results of CoTs to cater for phrasing text output.

\textbf{ReAct~\cite{yao2022react}:} This chain leverages an LLM prompted with tasks and tools, where the tools are functions described for LLM to call. For example, searching among the training set can be a tool. In each step, the LLM decides whether it can reach an answer or needs to use a tool. If a tool is needed, the LLM specifies the tool name and argument for an executor to call. The executor runs the corresponding functions and feeds the output back into the LLM input for the next step. To highlight the importance of the agent's ability to choose from more than one tool, we included our VDB search function as one of the tools and added the medical abbreviation lookup tool, \emph{medialpy}\footnote{\url{https://pypi.org/project/medialpy/}}, as another. We used LangChain's default implementation to facilitate this process. The ReAct chain's capability to use heterogeneous sources makes it unique among the chains compared in this study.

\begin{table}[ht]
    \vspace{-0em} 
    \centering
    \caption{LLM Judge Interrogation Metrics (Scores range from 0 to 1)}
    \label{tab:interrogation_metric}
    \begin{tabular}{|m{1.3cm}|m{6.8cm}|}
    \hline
    \rowcolor[HTML]{EFEFEF} 
    \textbf{Metric} & \textbf{Description} \\
    \hline
    \textbf{Specificity} & Evaluates how precisely the question addresses the patient's described symptoms and medical context, ensuring the question targets clinically relevant details. \\
    \hline
    \textbf{Usefulness} & Assesses the diagnostic value of the question, focusing on its potential to elicit new or critical information that contributes to a better understanding of the patient's condition. \\
    \hline
    \textbf{Relevance} & Determines the alignment of the question with the patient’s symptoms, medical history, and overall context, ensuring it stays pertinent to the clinical scenario. \\
    \hline
    \textbf{Coverage} & Measures how comprehensively the question explores the available patient information, ensuring that no critical aspects of the patient's condition are overlooked. \\
    \hline
    \textbf{Fluency} & Evaluates the clarity and coherence of the question's formulation, ensuring it is expressed in clear and natural medical language appropriate for healthcare communication. \\
    \hline
    \end{tabular}
    \small
    \vspace{-0em} 
\end{table}

\subsection{Evaluation Framework}

Our evaluation framework employs three primary modules:

\begin{itemize}
    \item \textbf{Virtual Patient Simulation:} This module simulates the patient's first statement as the input to the LLM healthcare chains, generating responses based on the first generated question. The simulation mimics real-world patient interactions to provide valid input for the LLM healthcare chains. Specifically, \( VPS \) represents the virtual patient simulator function that, given the patient knowledge \( K \), generates the initial statement \( S_{init} \) (i.e., \( VPS:(.,K) \rightarrow Stmt \)). Similarly, \( VPS\) is used to generate a simulated answer from the LLM doctor's question \( Q \) based on the patient knowledge \( K \) (i.e., \( VPS:(Q, K) \rightarrow Ansr \)). The topic and factual content of the virtual patient are controlled by the patient knowledge \( K \), ensuring that the distribution of topics matches the test set. We prompt the virtual patient to generate only short conversations, providing information solely based on the given knowledge \( K \). The source of bias will be the LLM phrasing and inconsistency from \( K \).
    
    \item \textbf{LLM Judge Interrogation:} This metric evaluates the questions generated by the LLM healthcare chains based on specificity, usefulness, relevance, coverage, and fluency, as shown in Table~\ref{tab:interrogation_metric}. An LLM judge \( J \) assesses each question \( Q \) using the patient knowledge \( K \), providing scores according to these criteria. The interrogation metrics are calculated as follows:    
    \[
    \text{Interrogation Metrics}=J_K(Q_{i,j},K_i)
    \]
    where \(i\) is the index for the patient case and \(j\) is the index for the specific question generated for that patient case.
    
    \item \textbf{Summarization-Based Evaluation:} This metric evaluates the quality of information retrieved by the generated questions using techniques including Recall-Oriented Understudy for Gisting Evaluation (ROUGE) and Named Entity Recognition (NER)-based similarity. These tools assess how well the simulated patient answers encapsulate the entire patient case by comparing the generated responses to the comprehensive patient information from the test data.
\end{itemize}

The evaluation process involves the following steps:

\begin{enumerate}
    \item \textbf{Initialize Ground Truth Data ($GT$)}: 
    Collect comprehensive information about the patient's symptoms and background to create a dataset for evaluation.
    
    \item \textbf{Provide Patient Known Knowledge ($K$)}: 
    Share a subset of the ground truth $GT$ with the LLM healthcare chain, containing the partial knowledge the patient initially provides to the doctor.
    
    \item \textbf{Provide a Virtual Patient Statement ($Stmt = VPS(.,K)$)}: 
    Simulate the initial statement a patient would provide to the doctor using an LLM to generate this statement based on the partial knowledge, as patients typically do not provide all information at once. We retain the complete knowledge for evaluation purposes in later steps.
    
    \item \textbf{Generate Question ($Q$)}: 
    The LLM healthcare chain generates question $Q$ based on the virtual patient's first condition statement ($Stmt$).
    
    \item \textbf{Evaluate the Question ($Q$) using Judge ($J(Q, K)$)}:
    Evaluate the generated question using an LLM judge based on criteria such as relevance, specificity, and informativeness, as detailed in Table~\ref{tab:interrogation_metric}.

    \item \textbf{Generate Simulated Answer and Evaluate as Summary ($Ansr := VPS(Q, K)$)}: The LLM virtual patient generates an answer based on the initial question and the known patient information. We then evaluate how well this simulated answer encapsulates the patient's entire known knowledge ($K$) to determine the quality and completeness of the response.
\end{enumerate}

In this process, a simulated patient generates an initial statement (\(Stmt\)), which the LLM healthcare chain uses to propose questions. The evaluation system assesses these questions based on specific criteria, with access to additional ground truth data that the healthcare chains do not have. Although the healthcare chains can use tools and reflective processes, they are only given the \(Stmt\), and no other test data is provided. The tools are derived from training sets or other external sources, not from the test ground truth. The simulated input (\(K|Stmt\)), which contains less information than the full ground truth (\(GT\)), is shared. This ensures that the judges have complete knowledge of the test case, while the healthcare chains work within a fair and restricted evaluation process.

\subsection{Validation}
Our validation framework ensures the robustness and reliability of the evaluation by focusing on the dependencies between interrogation and summarization metrics. The analysis is conducted at an aggregate level, where the scores for interrogation and summarization metrics are averaged across all interactions for each LLM healthcare chain. This approach captures systematic patterns and relationships across chain implementations, minimizing the influence of isolated cases or outliers.

To quantify these dependencies, we employ normalized mutual information (NMI), which measures the shared dependency between two metrics. Mutual information \(I(X, Y)\) evaluates how much knowing one variable reduces uncertainty about the other:
\[
I(X, Y) = \sum_{x \in X} \sum_{y \in Y} p(x, y) \log \frac{p(x, y)}{p(x)p(y)},
\]
where \(p(x, y)\) represents the joint probability distribution of \(X\) and \(Y\), while \(p(x)\) and \(p(y)\) are their respective marginal probabilities. Mutual information captures nonlinear dependencies and provides a general measure of association between the metrics.

Since the magnitude of raw mutual information depends on the scales of \(X\) and \(Y\), we normalize it to ensure comparability across different metrics. The normalized mutual information (NMI) is computed as:
\[
NMI(X, Y) = \frac{I(X, Y)}{\min(H(X), H(Y))},
\]
where \(H(X)\) and \(H(Y)\) denote the entropies of \(X\) and \(Y\), respectively:
\[
H(X) = -\sum_{x \in X} p(x) \log p(x).
\]

NMI values range from 0 to 1, where 0 indicates no dependency, and 1 signifies a perfect association between \(X\) and \(Y\). For example, a high NMI between interrogation metrics such as Relevance and summarization metrics like ROUGE-L Recall indicates that questions rated as highly relevant are more likely to elicit comprehensive answers aligning with the ground truth. By focusing on NMI, we capture both linear and nonlinear relationships between the metrics, offering a holistic view of their dependencies.

The aggregate analysis calculates average scores for each interrogation and summarization metric across all interactions within a given LLM healthcare chain. This averaging process ensures that the dependencies identified reflect the overall performance of the chain rather than being skewed by individual instances. Figure~\ref{fig:mi_grouped_heatmap} illustrates the NMI matrix for the aggregated metrics, providing a comprehensive view of the relationships between interrogation and summarization metrics across the LLM healthcare chains.

The validation process is designed to systematically address the core research questions:
\begin{itemize}
    \item \textbf{RQ1: How well does the induced answer to the question summarize all the relevant information?} By analyzing the dependencies between summarization metrics (e.g., ROUGE, NER2NER) and interrogation metrics, we assess whether high-quality questions generate responses that effectively encapsulate the ground truth.
    \item \textbf{RQ2: How correlated are the LLM judge interrogation metrics with the quality of the induced answers?} By leveraging NMI to quantify dependencies, we determine whether chains generating better-rated questions (e.g., higher Relevance or Usefulness) also elicit more comprehensive and informative answers.
\end{itemize}

By conducting the analysis at the aggregate level, we ensure that the framework captures general trends and patterns across different chains, reducing the noise introduced by outlier interactions. Furthermore, the use of normalized mutual information ensures that the evaluation is consistent and comparable across diverse metrics. The structure of the validation framework, which separates question generation from evaluation and prevents direct access to test ground truth, ensures fairness, avoids data leakage, and strengthens the credibility of the evaluation process. To facilitate better visualization and comparison across metrics with varying scales, scores were normalized to a common range through a linear rescaling process. This ensures consistent representation of the performance metrics while preserving the relative differences between them.

\section{Experiments}
\label{sec:experiments}

\subsection{Experiment Setup}
To ensure the feasibility of our approach for local implementations in clinics and hospitals, all components of the LLM health chains are based on open-source models where possible. The primary LLM used for the health chain processes is Mixtral\footnote{\url{https://huggingface.co/mistralai/Mixtral-8x7B-v0.1}}, served on Groq\footnote{\url{https://wow.groq.com/}}. For evaluating the generated questions, GPT-3.5-turbo and GPT-4 (via OpenAI API\footnote{\url{https://platform.openai.com/docs/models/gpt-4}}), along with Claude-3-Opus\footnote{\url{https://claude.ai/}}, are used as LLM judges. These models ensure robustness in evaluation and enable comparative analysis across different judging capabilities.

For chains using Chain of Thought Self-Consistency (CoT-SC) processes, a temperature of 0.3 is used to introduce slight variability, while a temperature of 0 is applied for all other models to ensure deterministic outputs. The NER component leverages an open-source biomedical NER model~\cite{raza2022large} to extract medical entities, which are then compared using set-based similarity metrics.

For retrieval tasks, we employ FAISS~\cite{douze2024faiss} with its default settings as the vector database (VDB) indexing package. Text embeddings are generated using the MiniLM\footnote{\url{https://huggingface.co/sentence-transformers/all-MiniLM-L6-v2}} model. Data preprocessing—including the extraction of patient known knowledge (\( K \)), virtual patient simulation, and question evaluation by the LLM judge—is conducted using Claude-3-Opus, GPT-3.5-turbo, and GPT-4 to ensure consistency and coverage in the evaluation pipeline.

The experiments are conducted on a dataset comprising 128 test cases (64 from each of the two datasets). We benchmark seven different LLM health chains, generating a total of 796 patient status-doctor statement pairs. For comparative analysis, the data is grouped by LLM chain type, and evaluation results are summarized for each model used as a judge. This includes mean scores from interrogation metrics such as specificity, usefulness, relevance, coverage, and fluency (see Table~\ref{tab:llm_judge}). The results are further validated by analyzing correlations between the LLM judges and traditional metrics like ROUGE and NER2NER.

\subsection{Evaluation Metrics Implementation}

To robustly evaluate the ability of LLM healthcare chains to gather useful patient information through effective questioning, we employ three complementary approaches: LLM judge interrogation, NER-based comparison, and ROUGE-based summarization evaluation.

The \textbf{LLM judge interrogation} assesses question quality using five key metrics: \textit{Relevance}, \textit{Usefulness}, \textit{Specificity}, \textit{Coverage}, and \textit{Fluency}. These metrics evaluate how well questions target essential patient information, their potential to uncover diagnostic insights, alignment with patient context, comprehensiveness, and linguistic clarity. Scores range from 0 to 1, with higher scores indicating better performance. Evaluations are conducted using advanced models such as GPT-3.5, GPT-4, and Claude-3-Opus to ensure robustness and diversity.

The \textbf{NER-based comparison} leverages a biomedical NER model~\cite{raza2022large} trained to identify key medical entities such as symptoms (e.g., physical signs or complaints) and medications (e.g., treatments or prescriptions). It evaluates how effectively the generated questions elicit responses containing relevant medical information by comparing the entities extracted from answers against the ground truth patient data.

The \textbf{ROUGE-based summarization evaluation} measures how closely the information elicited by the LLM-generated questions aligns with the comprehensive ground truth patient information. ROUGE-1 examines word-level overlap for basic coverage, ROUGE-2 assesses two-word phrase overlap to capture more complex patterns, and ROUGE-L evaluates the longest matching subsequences to gauge overall flow and completeness.

The gold standard for all evaluations is derived from the original patient notes in the test set, encompassing a complete range of symptoms, medical history, and contextual details. This comprehensive ground truth ensures that our framework rigorously measures both the breadth and depth of patient information elicited by the questions. Together, the LLM judge, NER, and ROUGE metrics quantify the relationship between question quality and the informativeness of elicited answers, providing a holistic assessment of the LLM healthcare chains' effectiveness.

\subsection{Results and Analysis}

\begin{table*}[t]
    \centering
    \caption{Performance of LLM Healthcare Chains Evaluated by LLM Judge Metrics}
    \label{tab:llm_judge}
    \scriptsize
    \setlength{\tabcolsep}{3pt}
    \resizebox{\textwidth}{!}{%
    \begin{tabular}{l|ccccc|ccccc|ccccc}
    \toprule
    LLM Chain & \multicolumn{5}{c}{GPT-4o} & \multicolumn{5}{c}{GPT-3.5} & \multicolumn{5}{c}{Claude} \\
    \cmidrule(lr){2-6} \cmidrule(lr){7-11} \cmidrule(lr){12-16}
    & Specificity & Usefulness & Relevance & Coverage & Fluency
    & Specificity & Usefulness & Relevance & Coverage & Fluency
    & Specificity & Usefulness & Relevance & Coverage & Fluency \\
    \midrule
    \rowcolor{green!15} RAG & 7.38 & 8.27 & 8.19 & 4.54 & 9.36 & 7.18 & 7.99 & 8.98 & 5.18 & 9.08 & 7.31 & 7.96 & 8.88 & 4.89 & 9.06 \\
    RAG\_default & 6.80 & 7.95 & 7.89 & 4.56 & 9.01 & 6.30 & 7.20 & 8.59 & 4.94 & 9.00 & 5.93 & 6.07 & 6.69 & 5.73 & 8.24 \\
    \rowcolor{yellow!25} RAG\_reflection & 7.98 & 8.82 & 8.72 & 5.39 & 9.47 & 7.84 & 8.53 & 9.18 & 5.77 & 9.23 & 7.88 & 8.38 & 9.28 & 5.64 & 9.17 \\
    \rowcolor{blue!15} RAG\_reflection\_cot & 7.68 & 8.54 & 8.43 & 5.11 & 9.33 & 7.50 & 8.21 & 9.05 & 5.53 & 9.13 & 7.57 & 8.13 & 9.07 & 5.35 & 9.12 \\
    RAG\_reflection\_cot\_sc & 7.28 & 8.12 & 8.03 & 4.90 & 9.12 & 7.02 & 7.89 & 8.82 & 5.24 & 9.00 & 7.12 & 7.86 & 8.79 & 5.09 & 8.98 \\
    ReAct & 5.52 & 5.59 & 5.46 & 3.63 & 8.44 & 4.59 & 5.75 & 5.68 & 4.34 & 7.92 & 3.65 & 4.70 & 3.83 & 2.29 & 7.84 \\
    Hardcoded & 5.30 & 5.02 & 4.63 & 3.69 & 6.70 & 5.26 & 5.18 & 4.83 & 4.38 & 6.80 & 4.30 & 5.31 & 5.41 & 3.70 & 6.57 \\
    \bottomrule
    \end{tabular}
    }
\end{table*}

\begin{figure*}[ht]
    \centering
    \includegraphics[width=1\textwidth]{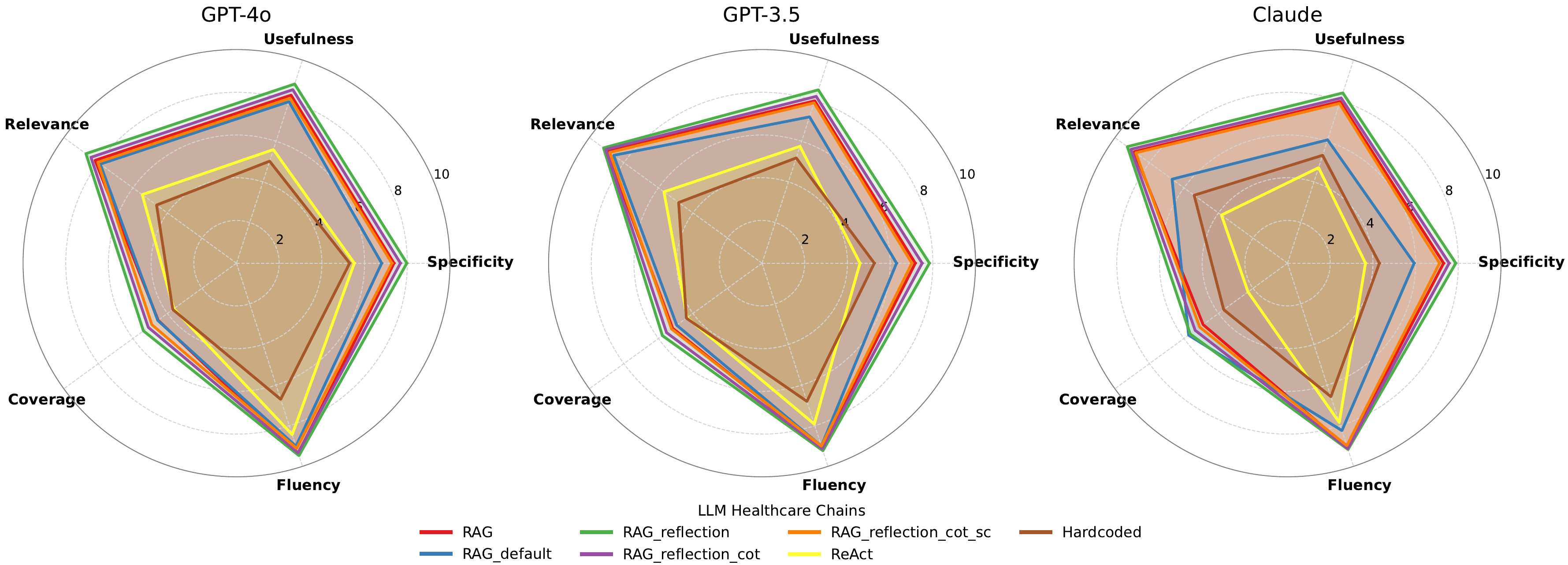}
    \caption{LLM judge interrogation metrics: radar plots}
    \label{fig:llm_judge_rader}
\end{figure*}

\begin{table}[t]
    \centering
    \begin{subtable}{\columnwidth}
        \centering
        \caption{NER Metrics Evaluation}
        \label{tab:full_rescaled_ner_results_highlighted}
        \resizebox{\columnwidth}{!}{%
        \begin{tabular}{llrrr}
        \toprule
        Model & Chain & NER2NER\_sym & NER2NER\_med & NER2NER\_total \\
        \midrule
        \multirow{7}{*}{GPT-4o}
        & \cellcolor{green!15} RAG & \cellcolor{green!15} 0.37 & \cellcolor{green!15} 0.27 & \cellcolor{green!15} 0.47 \\
        & RAG\_default & 0.20 & 0.15 & 0.34 \\
        & \cellcolor{yellow!25} RAG\_reflection & \cellcolor{yellow!25} 0.52 & \cellcolor{yellow!25} 0.42 & \cellcolor{yellow!25} 0.65 \\
        & \cellcolor{blue!15} RAG\_reflection\_cot & \cellcolor{blue!15} 0.44 & \cellcolor{blue!15} 0.33 & \cellcolor{blue!15} 0.54 \\
        & RAG\_reflection\_cot\_sc & 0.29 & 0.23 & 0.40 \\
        & ReAct & 0.10 & 0.09 & 0.15 \\
        & Hardcoded & 0.08 & 0.07 & 0.11 \\
        \midrule
        \multirow{7}{*}{GPT-3.5}
        & \cellcolor{green!15} RAG & \cellcolor{green!15} 0.32 & \cellcolor{green!15} 0.24 & \cellcolor{green!15} 0.42 \\
        & RAG\_default & 0.18 & 0.14 & 0.30 \\
        & \cellcolor{yellow!25} RAG\_reflection & \cellcolor{yellow!25} 0.45 & \cellcolor{yellow!25} 0.35 & \cellcolor{yellow!25} 0.57 \\
        & \cellcolor{blue!15} RAG\_reflection\_cot & \cellcolor{blue!15} 0.38 & \cellcolor{blue!15} 0.28 & \cellcolor{blue!15} 0.49 \\
        & RAG\_reflection\_cot\_sc & 0.26 & 0.20 & 0.34 \\
        & ReAct & 0.12 & 0.10 & 0.17 \\
        & Hardcoded & 0.09 & 0.07 & 0.13 \\
        \midrule
        \multirow{7}{*}{Claude}
        & \cellcolor{green!15} RAG & \cellcolor{green!15} 0.34 & \cellcolor{green!15} 0.25 & \cellcolor{green!15} 0.45 \\
        & RAG\_default & 0.22 & 0.16 & 0.36 \\
        & \cellcolor{yellow!25} RAG\_reflection & \cellcolor{yellow!25} 0.50 & \cellcolor{yellow!25} 0.40 & \cellcolor{yellow!25} 0.63 \\
        & \cellcolor{blue!15} RAG\_reflection\_cot & \cellcolor{blue!15} 0.42 & \cellcolor{blue!15} 0.33 & \cellcolor{blue!15} 0.54 \\
        & RAG\_reflection\_cot\_sc & 0.31 & 0.24 & 0.41 \\
        & ReAct & 0.13 & 0.10 & 0.16 \\
        & Hardcoded & 0.10 & 0.08 & 0.14 \\
        \bottomrule
        \end{tabular}%
        }
    \end{subtable}

    \vspace{1em}
    
    \begin{subtable}{\columnwidth}
        \centering
        \caption{ROUGE Metrics Evaluation}
        \label{tab:full_rescaled_rouge_results_highlighted}
        \resizebox{\columnwidth}{!}{%
        \begin{tabular}{llrrrrrrr}
        \toprule
        Model & Chain & rouge-1\_f & rouge-2\_r & rouge-2\_p & rouge-2\_f & rouge-l\_r & rouge-l\_p & rouge-l\_f \\
        \midrule
        \multirow{7}{*}{GPT-4o}
        & \cellcolor{green!15} RAG & \cellcolor{green!15} 0.66 & \cellcolor{green!15} 0.44 & \cellcolor{green!15} 0.40 & \cellcolor{green!15} 0.47 & \cellcolor{green!15} 0.62 & \cellcolor{green!15} 0.50 & \cellcolor{green!15} 0.54 \\
        & RAG\_default & 0.60 & 0.37 & 0.34 & 0.41 & 0.55 & 0.44 & 0.47 \\
        & \cellcolor{yellow!25} RAG\_reflection & \cellcolor{yellow!25} 0.75 & \cellcolor{yellow!25} 0.52 & \cellcolor{yellow!25} 0.48 & \cellcolor{yellow!25} 0.57 & \cellcolor{yellow!25} 0.70 & \cellcolor{yellow!25} 0.58 & \cellcolor{yellow!25} 0.62 \\
        & \cellcolor{blue!15} RAG\_reflection\_cot & \cellcolor{blue!15} 0.68 & \cellcolor{blue!15} 0.45 & \cellcolor{blue!15} 0.42 & \cellcolor{blue!15} 0.49 & \cellcolor{blue!15} 0.64 & \cellcolor{blue!15} 0.53 & \cellcolor{blue!15} 0.56 \\
        & RAG\_reflection\_cot\_sc & 0.48 & 0.32 & 0.30 & 0.34 & 0.45 & 0.36 & 0.39 \\
        & ReAct & 0.28 & 0.18 & 0.16 & 0.20 & 0.26 & 0.21 & 0.23 \\
        & Hardcoded & 0.17 & 0.12 & 0.10 & 0.14 & 0.15 & 0.12 & 0.13 \\
        \midrule
        \multirow{7}{*}{GPT-3.5}
        & \cellcolor{green!15} RAG & \cellcolor{green!15} 0.70 & \cellcolor{green!15} 0.46 & \cellcolor{green!15} 0.42 & \cellcolor{green!15} 0.50 & \cellcolor{green!15} 0.65 & \cellcolor{green!15} 0.54 & \cellcolor{green!15} 0.57 \\
        & RAG\_default & 0.57 & 0.38 & 0.35 & 0.43 & 0.53 & 0.44 & 0.47 \\
        & \cellcolor{yellow!25} RAG\_reflection & \cellcolor{yellow!25} 0.78 & \cellcolor{yellow!25} 0.55 & \cellcolor{yellow!25} 0.51 & \cellcolor{yellow!25} 0.60 & \cellcolor{yellow!25} 0.72 & \cellcolor{yellow!25} 0.61 & \cellcolor{yellow!25} 0.65 \\
        & \cellcolor{blue!15} RAG\_reflection\_cot & \cellcolor{blue!15} 0.71 & \cellcolor{blue!15} 0.48 & \cellcolor{blue!15} 0.44 & \cellcolor{blue!15} 0.53 & \cellcolor{blue!15} 0.67 & \cellcolor{blue!15} 0.56 & \cellcolor{blue!15} 0.60 \\
        & RAG\_reflection\_cot\_sc & 0.49 & 0.33 & 0.31 & 0.38 & 0.46 & 0.37 & 0.40 \\
        & ReAct & 0.29 & 0.19 & 0.17 & 0.21 & 0.27 & 0.22 & 0.24 \\
        & Hardcoded & 0.18 & 0.13 & 0.11 & 0.15 & 0.16 & 0.13 & 0.14 \\
        \midrule
        \multirow{7}{*}{Claude}
        & \cellcolor{green!15} RAG & \cellcolor{green!15} 0.64 & \cellcolor{green!15} 0.43 & \cellcolor{green!15} 0.39 & \cellcolor{green!15} 0.46 & \cellcolor{green!15} 0.60 & \cellcolor{green!15} 0.48 & \cellcolor{green!15} 0.52 \\
        & RAG\_default & 0.55 & 0.36 & 0.33 & 0.39 & 0.51 & 0.42 & 0.45 \\
        & \cellcolor{yellow!25} RAG\_reflection & \cellcolor{yellow!25} 0.72 & \cellcolor{yellow!25} 0.51 & \cellcolor{yellow!25} 0.47 & \cellcolor{yellow!25} 0.55 & \cellcolor{yellow!25} 0.67 & \cellcolor{yellow!25} 0.55 & \cellcolor{yellow!25} 0.59 \\
        & \cellcolor{blue!15} RAG\_reflection\_cot & \cellcolor{blue!15} 0.65 & \cellcolor{blue!15} 0.44 & \cellcolor{blue!15} 0.40 & \cellcolor{blue!15} 0.48 & \cellcolor{blue!15} 0.61 & \cellcolor{blue!15} 0.50 & \cellcolor{blue!15} 0.54 \\
        & RAG\_reflection\_cot\_sc & 0.46 & 0.31 & 0.29 & 0.33 & 0.43 & 0.34 & 0.37 \\
        & ReAct & 0.26 & 0.17 & 0.15 & 0.19 & 0.25 & 0.20 & 0.22 \\
        & Hardcoded & 0.15 & 0.11 & 0.09 & 0.13 & 0.14 & 0.11 & 0.12 \\
        \bottomrule
        \end{tabular}%
        }
    \end{subtable}
    
    \begin{minipage}{\columnwidth}
        \small
        \vspace{0.3em}
        \textcolor{yellow}{Yellow}: Best chain (\textbf{Reflection}),
        \textcolor{blue}{Blue}: Second-best (\textbf{Reflection\_CoT}),
        \textcolor{green}{Green}: Third-best (\textbf{RAG})
    \end{minipage}
    \caption{(a) NER Metrics: Evaluates entity recognition to assess the information elicited by questions. (b) ROUGE Metrics: Analyzes response comprehensiveness and precision. Together, these metrics demonstrate how effective questions drive informative answers in LLM healthcare chains.}

    \label{tab:QA_quality_comparison}
\end{table}

This section presents the evaluation results of the LLM healthcare chains, with explicit connections to the research questions \textbf{RQ1} and \textbf{RQ2}, which were established to validate the effectiveness of our proposed evaluation framework.

The results in Table~\ref{tab:QA_quality_comparison} strongly validate the effectiveness of our framework in addressing \textbf{RQ1}, which assesses how well the induced answers summarize all relevant information. Advanced chains such as RAG\_Reflection and RAG\_Reflection\_CoT consistently outperform simpler baselines like hardcoded and ReAct across all evaluation metrics, including both NER-based and ROUGE scores. For example, RAG\_Reflection achieves the highest NER2NER\_total scores (\textbf{0.65} for GPT-4o, \textbf{0.57} for GPT-3.5, and \textbf{0.63} for Claude), indicating its ability to extract critical medical entities effectively. Similarly, its superior ROUGE-L F scores (\textbf{0.62}, \textbf{0.65}, and \textbf{0.59}, respectively) highlight its capacity to generate responses that comprehensively align with ground truth patient information. Importantly, the consistency of these results across diverse LLM judges---GPT-4o, GPT-3.5, and Claude---demonstrates the robustness and model-agnostic nature of our framework. Despite the intrinsic differences in these LLMs, our evaluation framework consistently identifies similar performance trends, reinforcing its reliability as a tool for assessing questioning capabilities. Moreover, the significant performance gap between advanced chains and simpler baselines underscores the importance of well-crafted questions in eliciting informative responses. For instance, the hardcoded chain’s lower ROUGE-L F scores (\textbf{0.13} across judges) reveal its inability to generate effective interactions, affirming that high-quality questions play a pivotal role in summarizing patient-relevant information. These findings not only validate the utility of our framework but also emphasize the critical link between questioning capabilities and effective information elicitation, addressing \textbf{RQ1} comprehensively.

The results in Table~\ref{tab:llm_judge} and Fig.~\ref{fig:llm_judge_rader} provide comprehensive insights into \textbf{RQ2}, which explores the relationship between question quality and answer quality. Advanced LLM healthcare chains such as RAG\_reflection and RAG\_reflection\_CoT demonstrate significantly better performance across interrogation metrics compared to simpler methods like ReAct and hardcoded chains. For instance, RAG\_reflection achieves the highest scores in specificity (7.98 for GPT-4o, 7.84 for GPT-3.5, and 7.88 for Claude), reflecting its ability to generate precise and diagnostically focused questions. Similarly, RAG\_reflection\_CoT shows strong fluency scores (9.33 for GPT-4o, 9.13 for GPT-3.5, and 9.12 for Claude), benefiting from iterative chain-of-thought reasoning that enhances clarity and coherence. Coverage also improves with advanced chains; RAG\_reflection achieves a score of 5.39 under GPT-4o, significantly higher than the hardcoded chain’s 3.69. These trends indicate that reflection and iterative reasoning mechanisms enable chains to generate contextually rich and useful questions, leading to higher-quality responses. Conversely, the hardcoded and ReAct chains consistently lag across all metrics, with ReAct scoring as low as 5.46 in relevance and 3.63 in coverage under GPT-4o, underscoring their inability to adapt dynamically to patient contexts. These results confirm that advanced techniques not only improve question quality but also drive more comprehensive patient information elicitation.

Fig.~\ref{fig:llm_judge_rader} further illustrates the robustness and consistency of the proposed framework across different LLM judge models, including GPT-4o, GPT-3.5, and Claude. The radar plots highlight that advanced chains like RAG\_reflection and RAG\_reflection\_CoT consistently cover larger, more balanced areas, excelling across metrics such as specificity, relevance, and fluency. For example, the fluency of RAG\_reflection remains high across all judges (9.47 for GPT-4o, 9.23 for GPT-3.5, and 9.17 for Claude), suggesting that its question-generation process is resilient to the underlying LLM used for evaluation. The ReAct and hardcoded chains, in contrast, show narrow and uneven distributions, reflecting their lack of versatility in generating effective questions. Importantly, the framework’s consistency across judges demonstrates its model-agnostic nature, reinforcing its reliability for assessing questioning capabilities regardless of the LLM base model. Furthermore, the alignment of interrogation metrics with traditional measures, as shown in Table~\ref{tab:QA_quality_comparison}, validates the utility of the framework in systematically capturing essential aspects of questioning quality. For instance, RAG\_reflection achieves the highest ROUGE-L scores across all judges (0.70 for GPT-4o, 0.72 for GPT-3.5, and 0.67 for Claude), closely correlating with its high relevance and specificity metrics. These findings substantiate the central hypothesis that well-crafted questions lead to better patient responses, demonstrating the framework’s potential for evaluating and optimizing questioning capabilities systematically.

\begin{figure}[!ht]
    \centering
    \includegraphics[width=\columnwidth]{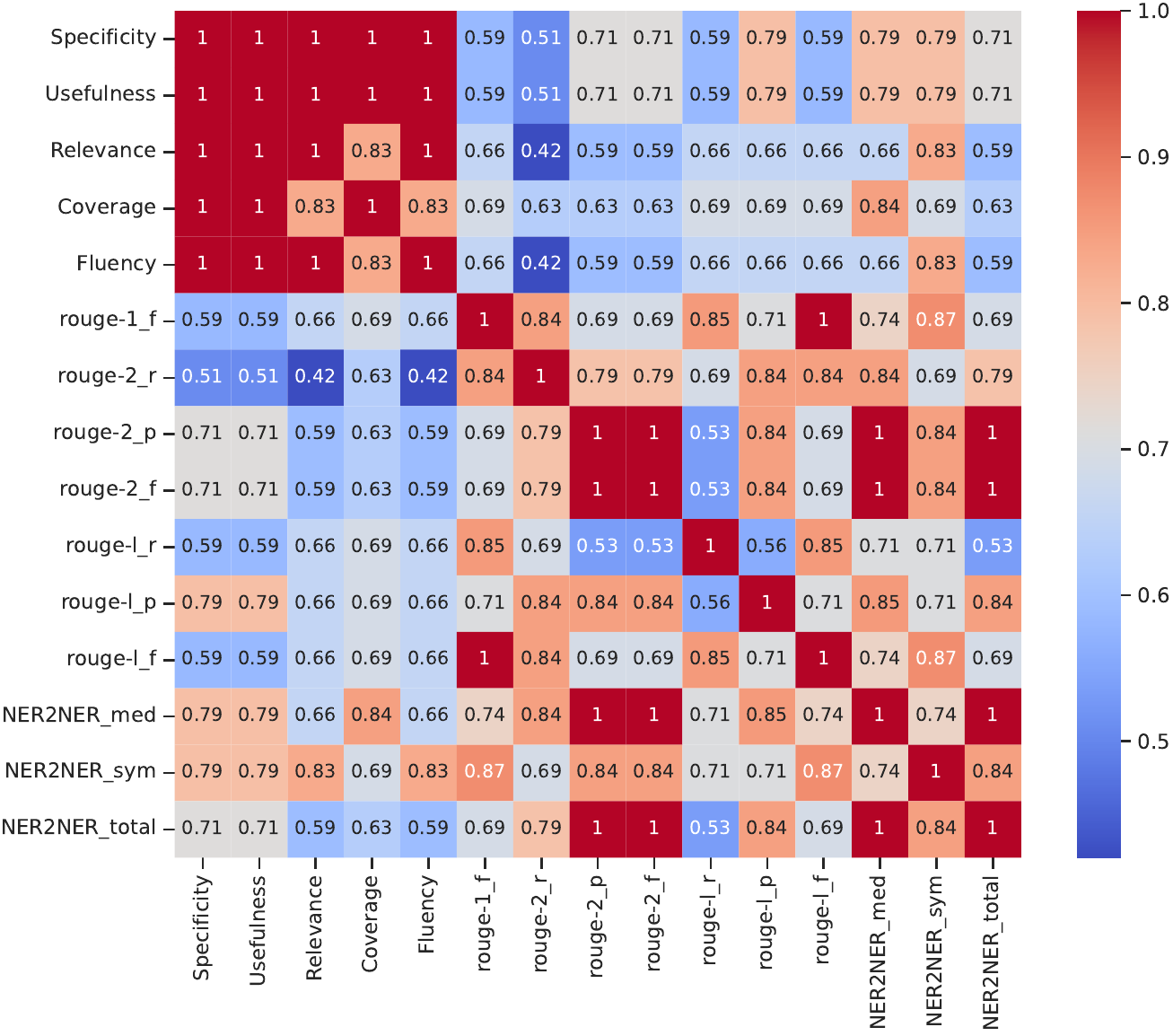} 
    \caption{Aggregated within specific LLM healthcare chains: Mutual information analysis - correlation between question quality and answer quality across all metrics.}
    \label{fig:mi_grouped_heatmap}
\end{figure}

Figure~\ref{fig:mi_grouped_heatmap} illustrates the relationships between evaluation metrics, showcasing the dependencies between question quality and answer quality across aggregated LLM healthcare chain interactions. Using normalized mutual information (NMI), the analysis quantifies the degree to which metrics such as Relevance, Usefulness, and traditional summarization measures like ROUGE and NER2NER align in evaluating chain performance.

The matrix reveals strong dependencies between Relevance and ROUGE-L Recall (\(NMI = 0.85\)) as well as Usefulness and NER2NER\_total (\(NMI = 0.84\)). These findings demonstrate that questions rated highly for their relevance and diagnostic utility are likely to elicit responses that closely align with the ground truth. This underscores the ability of well-targeted questions to uncover critical patient information effectively, reinforcing the connection between LLM-specific interrogation metrics and traditional summarization measures.

In contrast, weaker dependencies are observed between Fluency and NER2NER\_med (\(NMI = 0.66\)), as well as Coverage and ROUGE-2 Precision (\(NMI = 0.53\)). While fluency ensures linguistic clarity and coherence, these results indicate that it does not necessarily correlate with the extraction of detailed medical entities or precise information. Similarly, the modest correlation between Coverage and ROUGE-2 Precision suggests that broad or exhaustive questions do not always translate into precise and specific responses, highlighting potential trade-offs between general coverage and specificity.

Specificity, on the other hand, shows a stronger alignment with summarization metrics such as ROUGE-1 Recall (\(NMI = 0.79\)) and NER2NER\_total (\(NMI = 0.71\)). This highlights the critical role of specificity in driving targeted and informative interactions. Specific questions are more likely to elicit detailed and relevant answers that span the full spectrum of patient-relevant information, emphasizing their importance in diagnostic contexts.

Overall, the strong correlations between interrogation metrics like Relevance and Usefulness with summarization measures such as ROUGE-L validate the robustness of the evaluation framework. These results confirm that high-quality, targeted questions consistently lead to more comprehensive and accurate responses. Conversely, the weaker correlations for Fluency emphasize that while clarity is necessary, it does not guarantee informativeness or precision, pointing to the need for a balanced approach in designing LLM healthcare chains. These insights affirm the framework’s ability to capture nuanced dependencies and provide a holistic assessment of question-answering performance.

\section{Discussion}

\subsection{Sources of Bias and Correction}

Our evaluation framework considers potential biases in both the sampling process and the construction of \(Stmt\) and \(GT\).

\subsubsection{Pre-judgement Errors}

Randomized train-test splits and sampling from both ChatDoctor and MTS-Dialog datasets can lead to distributional biases. Re-weighting patient types can help mitigate this issue. Errors may arise from the LLM's data cleaning process due to hallucinations, or from patient case data that does not fully capture essential diagnostic information. We assume the patient data is comprehensive and accurately reflects real-world doctor-patient interactions. It may be possible to reduce this bias through conditioning on patient profiling and matching them with simulation promptings. Bias can also occur in transitioning from \(K\) to \(Stmt\), as patient statements may vary. Addressing this as a sample bias problem with fixed conversation starters can help correct this.

\subsubsection{Judgement Errors}

Interrogation biases stem from LLM prompting and judgement errors, while summarization metrics biases arise from LLM-simulated answers. Mitigating these biases involves diversifying error sources by using LLM capabilities differently: one for scoring directly and another for extracting \(Q\) from \(K\). Effective evaluation should align with good diagnosis and ultimately improve patient outcomes.

\subsection{Evaluation Limitations}
A key limitation of our current framework is the reliance on LLMs for evaluation. Although we demonstrate consistency across different models (GPT-3.5, GPT-4, and Claude), this approach may overlook aspects of question quality that only human healthcare professionals could assess. Future work should incorporate practicing clinicians to evaluate generated questions and test them in real clinical settings. Such human-in-the-loop evaluation would provide insights into practical utility and effectiveness in actual medical conversations. Additionally, longitudinal studies with healthcare providers could help establish how these automated systems complement human medical expertise, particularly in assessing question appropriateness across specialties and alignment with diagnostic protocols.




\section{Conclusion}

This paper introduced HealthQ, a novel framework for evaluating the questioning capabilities of LLM healthcare chains, addressing the critical but underexplored role of question formulation in digital healthcare. By implementing advanced LLM chains, such as RAG and CoT, and leveraging an LLM judge alongside traditional NLP metrics like ROUGE and NER-based evaluations, we demonstrated the importance of well-crafted questions in eliciting informative patient responses. Validated across multiple LLM judge models and datasets, our results showed consistent performance, highlighting the framework's robustness and potential for model-agnostic evaluation. HealthQ provides a comprehensive methodology for linking question quality to answer informativeness, setting a benchmark for advancing the effectiveness of LLMs in patient care. Future work will focus on expanding to multi-turn interactions, refining evaluation metrics, and exploring broader healthcare applications.

\bibliographystyle{IEEEtran}
\bibliography{ref}
\end{document}